\documentclass[runningheads,anonymous]{llncs}
\usepackage[T1]{fontenc}
\usepackage{graphicx}
\usepackage{wrapfig}

\usepackage{pifont}
\usepackage{amsmath}
\usepackage{booktabs}
\usepackage{multirow}
\usepackage{subcaption}

\usepackage{xcolor} 

\begin{document}
\title{SELAUR: Self Evolving LLM Agent via Uncertainty-aware Rewards}

%
%

\author{Dengjia Zhang\inst{1} \and
Xiaoou Liu\inst{2} \and
Lu Cheng\inst{3} \and 
Yaqing Wang\inst{4} \and 
Kenton Murray\inst{1} \and
Hua Wei\inst{2}
}
\authorrunning{D. Zhang et al.}
\institute{Johns Hopkins University, Baltimore MD, USA 
\email{\{dzhang98,kenton\}@jhu.edu}\\ \and
Arizona State University, Tempe AZ, USA
\email{\{xiaoouli,hua.wei\}@asu.edu} \\ \and 
University of Illinois Chicago, Chicago IL, USA 
\email{lucheng@uic.edu} \and 
Purdue University, West Lafayette IN, USA
\email{wang5075@purdue.edu}
}

\maketitle              
\begin{abstract}

Large language models (LLMs) are increasingly deployed as multi-step decision-making agents, where effective reward design is essential for guiding learning. Although recent work explores various forms of reward shaping and step-level credit assignment, a key signal remains largely overlooked: the intrinsic uncertainty of LLMs. Uncertainty reflects model confidence, reveals where exploration is needed, and offers valuable learning cues even in failed trajectories.
We introduce SELAUR: Self Evolving LLM Agent via Uncertainty-aware Rewards, a reinforcement learning framework that incorporates uncertainty directly into the reward design. SELAUR integrates entropy-, least-confidence-, and margin-based metrics into a combined token-level uncertainty estimate, providing dense confidence-aligned supervision, and employs a failure-aware reward reshaping mechanism that injects these uncertainty signals into step- and trajectory-level rewards to improve exploration efficiency and learning stability.
Experiments on two benchmarks, ALFWorld and WebShop, show that our method consistently improves success rates over strong baselines. Ablation studies further demonstrate how uncertainty signals enhance exploration and robustness.\footnote{code is available: https://github.com/adoptedirelia/SELAUR-Self-Evolving-LLM-Agent-via-Uncertainty-aware-Rewards}
\keywords{Uncertainty  \and LLM Agent \and Reinforcement Learning}
\end{abstract}

\section{Introduction}

Large language models (LLMs) have rapidly evolved from solving single-turn question-answering tasks~\cite{brown2020language,roberts2020much} to functioning as interactive agents capable of multi-step reasoning and decision-making~\cite{plaat2025multi}.
In these agentic settings, an LLM no longer produces a single response but generates a sequence of actions and intermediate decisions, forming a complete interaction trajectory.
For instance, in \textit{ALFWorld} \cite{shridhar2020alfworld}, the agent must interact with a simulated environment through a series of actions to accomplish the final goal.
Reinforcement learning has become a common paradigm for optimizing such agents.
Conventional RL approaches typically assign rewards based solely on the final outcome \cite{kwon2024adaptive}, which drives the agent to maximize success but often overlooks the contribution of intermediate steps.
Recent work has introduced step-level credit assignment~\cite{feng2025group}, where each decision within a trajectory is evaluated according to its influence on the final outcome.
This fine-grained reward design allows the training process to capture more nuanced learning signals and improves policy optimization stability.

However, a major limitation of most existing approaches is the neglect of the \textit{intrinsic uncertainty} inherent in LLMs, even though such uncertainty can provide valuable information for guiding learning.
Uncertainty plays a critical role in training interactive agents for two main reasons.
First, it serves as a natural measure of model confidence, enabling the reward signal to reflect not only correctness but also confidence in each decision.
For example, in an information retrieval scenario, an agent may sequentially query multiple databases before reaching the correct result.
If the model exhibits high uncertainty at a particular step, this indicates an exploratory state where encouraging alternative strategies is beneficial.
Conversely, when the agent displays low uncertainty at a step, the reward signal should reinforce this reliable behavior to promote convergence.
Second, incorporating uncertainty allows agents to extract meaningful learning signals even from failed trajectories~\cite{zhu2025surprising}.
Even if the final result is incorrect, step-level uncertainty can reveal which decisions are uncertain or promising, providing useful feedback for refinement in subsequent updates.

In this work, we advance existing step-level reward mechanisms by advancing ~\textbf{S}elf \textbf{E}volving \textbf{L}LM \textbf{A}gent via \textbf{U}ncertainty-aware \textbf{R}ewards (SELAUR).
SELAUR integrates token-level uncertainty estimates into both step-level and trajectory-level evaluations, ensuring that the agent benefits from informative feedback across the entire learning process.
Specifically, we:
(1) explore multiple token-level uncertainty metrics (i.e., entropy, least confidence, and margin) and integrate them into a unified uncertainty measure;
(2) aggregate these signals into step- and trajectory-level confidence estimations; and
(3) incorporate these uncertainty-aware signals into both step-level and trajectory-level rewards.
This design enables the agent to leverage uncertainty not only to enhance exploration but also to extract value from failed experiences, leading to more balanced and stable policy learning.
We evaluate our method on two well-established interactive benchmarks, \textit{ALFWorld} and \textit{WebShop}~\cite{yao2022webshop}. The results show that by explicitly modeling failure signals in the reward function, our failure-aware reward reshaping method consistently improves task success rates compared to strong reinforcement learning baselines.
Our main contributions are summarized as follows:
(1) We introduce a unified uncertainty framework that integrates multiple token-level uncertainty measures, enabling a more accurate characterization of model confidence and providing dense, informative supervision derived directly from these confidence estimates.
(2) We introduce a failure-aware reward reshaping mechanism that integrates these uncertainty signals into both step- and trajectory-level rewards, improving exploration efficiency and stability.
(3) We empirically demonstrate that incorporating information from failed trajectories can provide valuable training signals.

\vspace{-5 pt}
\section{Related Work}

\textbf{Uncertainty Quantification in LLMs.} Uncertainty quantification (UQ) has been extensively studied in machine learning as a means of estimating prediction confidence~\cite{van2024uncertainty,shorinwa2025survey,chrysos2025quantify}. In natural language processing, token-level entropy and consistency-based measures~\cite{fomicheva2020uncertainty,malinin2020uncertainty} have been used for detecting unreliable outputs, improving robustness, and enabling selective prediction. For LLMs, uncertainty estimation has proven useful in tasks such as answer verification~\cite{kadavath2022language}, selective QA~\cite{kamath2020selective}, hallucination detection~\cite{manakul2023selfcheck}, continual learning \cite{zhou2025robust}, active RAG \cite{min2025quco}, LLMs self-improvement \cite{huang2025beyond}, and efficient reasoning \cite{su2025cp}. These studies underscore that uncertainty is a strong indicator of reliability.

In RL and decision-making, uncertainty has further been leveraged to drive exploration through mechanisms such as intrinsic motivation or optimism under uncertainty~\cite{yu2025reward,sukhija2025optimism,liu2024ovd}. However, few works have explicitly integrated uncertainty signals into the training of LLM agents~\cite{suri2025structured,zhao2024saup,xie2025unlocking}. Our approach bridges this gap by using token-level uncertainty to shape both step- and trajectory-level rewards, enabling agents to balance between exploiting confident knowledge and exploring uncertain but potentially valuable strategies.

\noindent\textbf{LLM Agents and Reinforcement Learning.}  Recent advancements have transformed large language models (LLMs) from single-turn predictors into general-purpose agents capable of multi-step reasoning, planning, and tool use~\cite{xi2025rise,wang2024survey,yehudai2025survey}. Benchmarks such as ALFWorld~\cite{shridhar2020alfworld} and WebShop~\cite{yao2022webshop} evaluate these abilities by requiring agents to execute sequences of actions in interactive environments. To improve performance, prior work has explored prompting techniques~\cite{kojima2022large}, tool augmentation~\cite{schick2023toolformer,shen2024exploring}, and trajectory optimization~\cite{song2024trial}. Methods like chain-of-thought prompting~\cite{wei2022chain} and planner–executor frameworks~\cite{he2025plan,sun2024pearl} enhance reasoning and task decomposition, yet they often depend on supervised trajectories that may not generalize broadly.

Reinforcement learning (RL) provides a complementary approach for training LLM agents toward complex objectives~\cite{aissi2025reinforcement,carta2023grounding}. While RLHF~\cite{christiano2017deep,ouyang2022training} has proven effective for optimizing final outputs, outcome-based rewards are typically sparse, offering limited guidance for intermediate reasoning steps~\cite{lyu2025exploring,ye2025beyond}. Step-level credit assignment methods~\cite{ng1999policy,jeon2020reward,huang2022inner} address this by distributing rewards across individual decisions, but they generally ignore model uncertainty. Recent agentic RL frameworks such as ReAct~\cite{yao2022react}, Voyager~\cite{wang2023voyager}, and Reflexion\cite{shinn2023reflexion} demonstrate the benefits of process-level feedback, yet they still rely heavily on explicit external or self-generated evaluations~\cite{zhang2025survey,qi2024webrl,liu2025survey}. In contrast, our proposed SELAUR framework leverages uncertainty-based intrinsic rewards as an internal learning signal. By transforming low-confidence or failed reasoning steps into structured feedback, SELAUR provides dense, informative supervision directly from model confidence estimates. This enables LLM agents to improve policy stability, recover from uncertain reasoning paths, and learn more effectively without relying on external environment rewards or reflection heuristics.

\section{Method}
\begin{figure*}[t]
    \centering
    \includegraphics[width=1\linewidth]{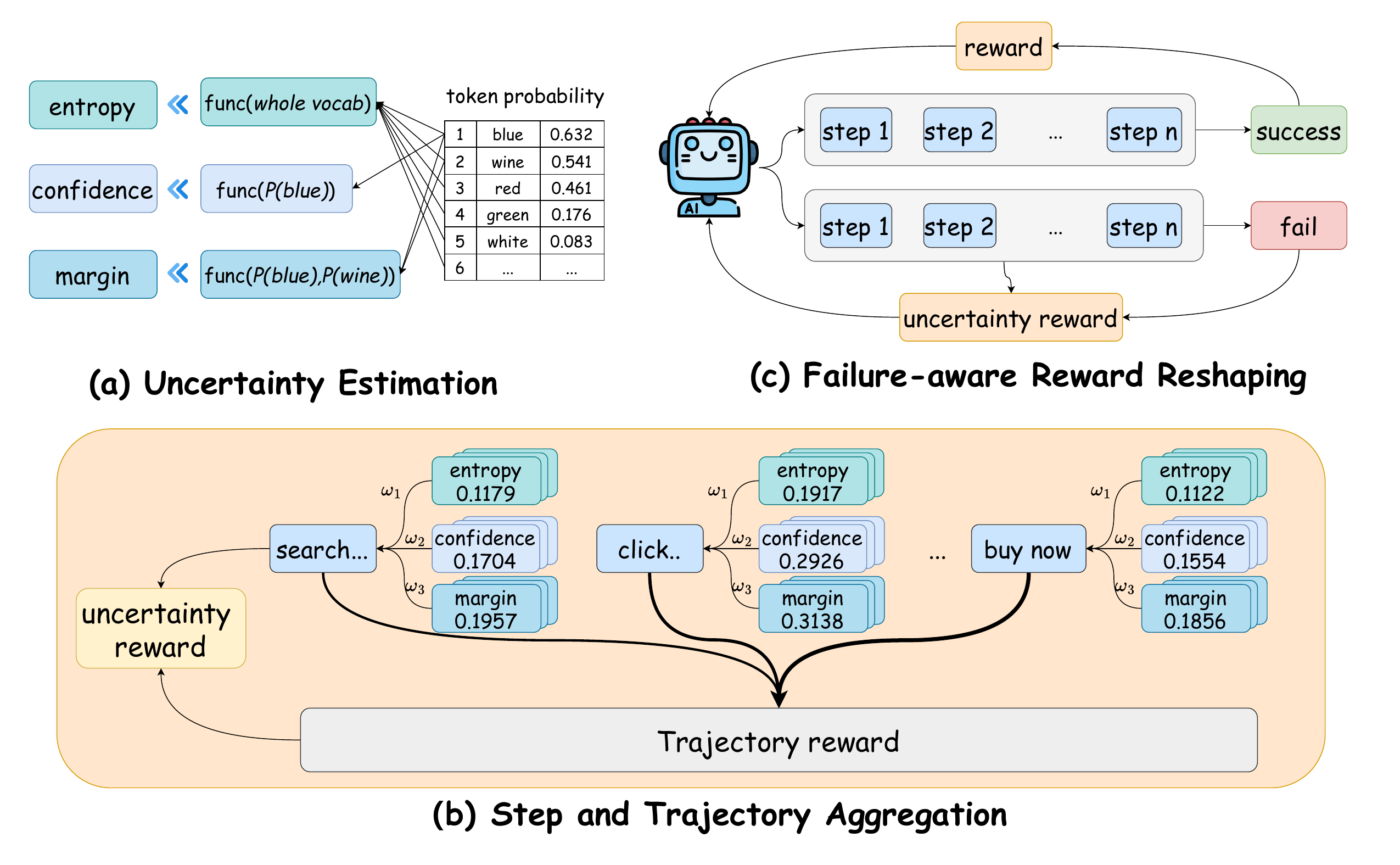}
    \caption{\small Overview of our method, SELAUR.
(a) \textbf{Uncertainty Estimation}: We employ three approaches to estimate model uncertainty: entropy, confidence, and margin, derived from token probability distributions.
(b) \textbf{Step and Trajectory Aggregation}: Throughout the model’s process, we compute two types of rewards that contribute to the uncertainty reward. At the step level, token-level uncertainties are aggregated. At the trajectory level, step-level uncertainties are combined with varying weights, where later steps receive higher weights in the overall trajectory reward.
(c) \textbf{Failure-aware Reward Reshaping}: For successful cases, the model is trained with the standard reward. For failed cases, we incorporate the uncertainty reward into the training loop to improve learning robustness.}
    \label{fig:model}
\vspace{-10 pt}
\end{figure*}

In this paper, we propose a novel reinforcement learning framework, ~\textbf{S}elf \textbf{E}volving \textbf{L}LM \textbf{A}gent via \textbf{U}ncertainty-aware \textbf{R}ewards (SELAUR), designed for LLM agents.

As illustrated in Figure \ref{fig:model}, SELAUR consists of three main modules:
(1) Uncertainty Estimation,
(2) Step and Trajectory Aggregation, and
(3) Failure-aware Reward Shaping.
This architecture ensures that failures are not discarded but instead transformed into learning opportunities, where uncertainty provides dense feedback about where the model hesitated or was less confident.

\subsection{Uncertainty Estimation}

A core component of SELAUR is quantifying the uncertainty of model at each decoding step.
Since different uncertainty measures capture distinct facets of predictive reliability, we combine three complementary metrics into a unified token-level uncertainty to capture uncertainty from multiple perspectives: global distributional spread (\textbf{entropy}), local decision confidence (\textbf{least confidence}), and comparative distinctiveness (\textbf{margin}).

\paragraph{Entropy.}  
Entropy~\cite{huang2024calibrating} measures the dispersion of the probability distribution over the vocabulary, reflecting how concentrated or spread out the model’s belief is.
A higher entropy indicates that the model distributes probability mass across many tokens, whereas a lower entropy implies strong confidence in a single token:
$u^{\text{ent}}_{t,j} = \frac{-\sum_c p_{t,c}\log p_{t,c}}{\log |\mathcal{V}|}$,
where $p_{t,c}$ is the predicted probability of token $c$ at step $t$, and $|\mathcal{V}|$ denotes the vocabulary size.  

\paragraph{Least confidence.}  

The least confidence measure~\cite{chen2012applying} assesses uncertainty using only the probability of the selected token. Lower predicted probability indicates higher uncertainty and higher predicted probability brings higher confidence:
$u^{\text{lc}}_{t,j} = 1 - p_{t,(1)}$,
where $p_{t,j}$ is the probability of chosen token at step $t$.  

\paragraph{Margin.}  

The margin-based uncertainty~\cite{yang2024maqa} captures uncertainty through the gap between the top two token probabilities. Smaller margins reflect greater ambiguity, whereas larger margins indicate stronger confidence: $
u^{\text{mar}}_{t,j} = \sigma\!\left(\frac{1-(p_{t,(1)}-p_{t,(2)})}{s}\right)$,
where $p_{t,(1)}$ and $p_{t,(2)}$ are the top-2 token probabilities, $s$ is a scaling factor, and $\sigma(\cdot)$ is the sigmoid function to normalize the score.  

\paragraph{Aggregated Uncertainty.}  
The final token-level uncertainty integrates the three measures with tunable weights: $ u_{t,j} = w_{\text{ent}} u^{\text{ent}}_{t,j} 
         + w_{\text{lc}} u^{\text{lc}}_{t,j} 
             + w_{\text{mar}} u^{\text{mar}}_{t,j}.$

This aggregation allows the framework to capture uncertainty from multiple perspectives with overall distributional spread, decision confidence, and distinctiveness.

\subsection{Step and Trajectory Aggregation}
Uncertainty at the token level provides granular feedback, but reinforcement learning requires signals at the step or trajectory scale.  
Therefore, we design the following hierarchical aggregation mechanisms as reward signals.  

\paragraph{Step-level aggregation.}  
At each step $t$, token uncertainties are aggregated by combining average values: $u^{\text{step}}_t = \frac{1}{j}\sum_j u_{t,j}.$

\paragraph{Trajectory-level aggregation.}  
To capture overall uncertainty across a trajectory $\tau$, we adopt an exponentially discounted aggregation, assigning greater weight to later steps, which often determine final success: $U(\tau) = \frac{\sum_{t=1}^{T} \lambda^{T-t} u^{\text{step}}_t}{\sum_{t=1}^{T} \lambda^{T-t}},$

where $\lambda \in (0,1]$ is the discount factor controlling the temporal emphasis. 

\subsection{Failure-aware Reward Shaping}

Standard RLVR formulations use binary outcome rewards (success/failure)~\cite{guo2025deepseek}, 
which risks discarding useful gradient signals from failed trajectories.  
SELAUR reshapes rewards by integrating uncertainty into per-step feedback, 
thus transforming failures into informative learning signals.

\paragraph{Step-wise shaping.}
When a trajectory fails, each step reward is augmented by its normalized uncertainty:
\begin{equation}
\tilde r_t^{\text{step}} = 
\begin{cases}
    \mathbf{1}[\text{fail}] \cdot w_t \, \hat u^{\text{step}}_t, & \text{if fail}, \\
    r_t, & \text{otherwise},
\end{cases}
\end{equation}
where $w_t$ is a step-dependent weight and $\hat u^{\text{step}}_t$ denotes the normalized step uncertainty. In our settings, $w_t$ is set to 0.95 to keep failure rewards informative but always smaller than the success rewards.

\paragraph{Trajectory-level shaping.}
Similarly, the trajectory reward is modified as:
\begin{equation}
\tilde r^{\text{traj}} = 
\begin{cases}
    U(\tau), & \text{if fail}, \\
    r, & \text{otherwise}.
\end{cases}
\end{equation}

This reward shaping mechanism converts failed experiences into dense and meaningful signals, 
encouraging the agent to revisit uncertain reasoning paths and refine its policy.  
In essence, SELAUR bridges the gap between sparse, outcome-based rewards and dense, token-level learning signals 
by aligning reward dynamics with model uncertainty.

\section{Experiments}

\subsection{Experimental Setup}
We evaluate our proposed method on two widely adopted interactive benchmarks: \textbf{ALFWorld}, which emphasizes generalization in instruction following, and \textbf{WebShop}, which focuses on goal-oriented decision-making in realistic e-commerce environments.
These benchmarks are complementary in nature: ALFWorld requires compositional reasoning across diverse household tasks, while WebShop challenges agents with open-ended product selection conditioned on natural goals. 

Experiments were conducted on \textbf{NVIDIA A100 and H100 GPUs}. For all experiments, we set the learning rate to $1 \times 10^{-6}$, the total number of training steps to 150, and the rollout size to 8. Considering the varying complexity of the tasks, we adapt the maximum interaction steps for each environment: specifically, the limit is set to 50 for ALFWorld and 15 for WebShop.

We compare our approach against several representative reinforcement learning baselines, including \textbf{PPO}~\cite{schulman2017proximal}, \textbf{RLOO}~\cite{ahmadian2024back}, \textbf{GRPO}~\cite{shao2024deepseekmath}, and \textbf{GiGPO}~\cite{feng2025group}.
For evaluation, following the existing work~\cite{wang2024soft}, we primarily report the \textbf{task success rate}, which directly measures whether the agent successfully accomplishes the tasks.
In addition, for the WebShop benchmark, we include the \textbf{task score}, which quantifies the alignment between user intent and the agent’s final product selection.

\begin{table}[t!]
\centering
\caption{\small Main experimental results on \textit{ALFWorld} and \textit{WebShop}.
For \textit{ALFWorld}, we report success rates (\%) on individual tasks and their overall average (All).
For \textit{WebShop}, we report both the task score and success rate (\%).\textbf{Bold numbers} mean the best performance, and \textit{italic numbers} mean the second-best. Overall, our method achieves the strongest performance across both benchmarks.}
\label{tab:main-results-big}
\resizebox{\textwidth}{!}{
\begin{tabular}{llccccccc|cc}
\toprule
\multirow{2}{*}{\textbf{Base Model}} & 
\multirow{2}{*}{\textbf{Method}} & 
\multicolumn{7}{c|}{\textbf{AlfWorld}} &
\multicolumn{2}{c}{\textbf{WebShop}} \\
\cmidrule(r){3-9} \cmidrule(r){10-11}

& &

\textbf{Pick} & \textbf{Look} & \textbf{Clean} & \textbf{Heat} & \textbf{Cool} & \textbf{Pick2} & \textbf{All} & 
\textbf{Score} & \textbf{Succ.} \\
\midrule
\multirow{6}{*}{Qwen2.5-1.5B}
& prompting & 0.1071&0.0000&0.0000&0.071&0.0000&0.0000&0.0312&0.2180&0.0429\\
& PPO &0.9142&0.3750&0.8333&0.8333&0.7368&0.7916&0.8046&0.7126 & 0.4921\\ 
& RLOO &0.9000&\textit{0.7000}&0.9200&0.7142&0.6875&0.5000&0.7812&0.8022&0.6394\\
& GRPO &0.6756  &0.5000&0.8000&\textit{0.9285}&0.6190&0.5263&0.6796&\textit{0.8359}&0.6718\\
& GiGPO & \textit{0.9143} & \textbf{0.7500} & \textbf{1.0000} &    0.8889 & \textbf{0.8421} & \textit{0.7916} & \textit{0.8828} & 0.8017 & \textit{0.6757} \\
& Ours(SELAUR)  & \textbf{0.9143} & 0.6250 & \textit{0.9583} & \textbf{1.0000} & 0.7894 & \textbf{0.8750} & \textbf{0.8906} & \textbf{0.8812} & \textbf{0.7656} \\
\midrule
\multirow{6}{*}{Qwen2.5-7B} 
& prompting &0.2800 &0.2500&0.1200&0.0000&0.0370&0.0000&0.1171&0.0288&0.0039\\
& PPO & 0.7241&0.7000&0.8800&0.4285&0.7500&0.6111&0.7109& 0.8689&0.7187\\
& RLOO &1.0000 &0.8000&1.0000&0.8571&0.6875&0.7778&0.8593&0.8768&0.7578\\
& GRPO &0.8275 &0.8000&0.9200&0.5714&0.6875&0.6111&0.7500&0.8184&0.7578\\
& GiGPO & \textit{1.0000} & \textit{1.0000} & \textit{1.0000} & \textit{0.9444} & \textit{0.7894} & \textit{0.9166} & \textit{0.9453} & \textit{0.8867} & \textit{0.7929} \\
& Ours(SELAUR)  & \textbf{1.0000} & \textbf{1.0000} & \textbf{1.0000} & \textbf{1.0000} & \textbf{0.9259} & \textbf{0.9259} & \textbf{0.9687} & \textbf{0.8935} & \textbf{0.7968} \\

\bottomrule
\end{tabular}}

\end{table}

\subsection{Main Results}

As shown in Table~\ref{tab:main-results-big}, the purely prompted agent performs poorly across both benchmarks, indicating that instruction-following ability alone is insufficient for complex interactive reasoning without reinforcement-based adaptation.
Among learning-based methods, \textbf{PPO} achieves competitive performance, especially on the smaller 1.5B model, but it suffers from lower efficiency because of its extra critic network.
\textbf{RLOO} and \textbf{GRPO} provide more balanced results, maintaining good performance while being computationally more efficient.
\textbf{GiGPO}, which introduces step-level reward modeling, further improves overall outcomes by enabling more fine-grained step reward throughout the trajectory.

Building upon these advances, our proposed method surpasses GiGPO by integrating uncertainty estimation into the RL process.
The inclusion of diverse uncertainty measures allows the agent to better identify and explore uncertain states, leading to more comprehensive policy improvement across different learning stages.
This uncertainty-aware exploration encourages the model to actively refine its decision-making process rather than overfitting to confident but suboptimal behaviors.
We provide a detailed qualitative discussion of this behavior in Section\ \ref{sec:case}.
\vspace{-10pt}

\begin{table}[h!]
\centering
\vspace{-5mm}
\caption{
    Each row evaluates a specific combination of metrics: \textbf{\ding{51}} indicates the inclusion of an uncertainty metric in the selection process, while \textbf{\ding{55}} denotes its exclusion. Results show that while individual or pairwise metrics provide partial gains, the full integration of Entropy, Least Confidence, and Margin ($\text{\ding{51}}$ across all) yields the optimal performance in both Task Score and Success Rate.
}
\label{tab:uncertainty}
\begin{tabular}{ccccc}
\toprule
\multicolumn{3}{c}{\textbf{Setting}} & 
\multirow{2}{*}{\textbf{Task Score}} & 
\multirow{2}{*}{\textbf{Success Rate}} \\
\cmidrule(r){1-3} \textbf{entropy} &\multicolumn{1}{c}{\textbf{least confidence}} & \textbf{margin} & &\\

\midrule
\ding{55} & \ding{55} & \ding{55} & 0.8017 & 0.6757 \\
\ding{51} & \ding{55} & \ding{55} & 0.8504 & 0.7265 \\
\ding{55} & \ding{51} & \ding{55} & 0.8465 & 0.7226 \\
\ding{55} & \ding{55} & \ding{51} & 0.7834 & 0.6328 \\
\ding{51} & \ding{51} & \ding{55} & 0.8263 & 0.6953 \\
\ding{55} & \ding{51} & \ding{51} & 0.8589 & 0.6562 \\
\ding{51} & \ding{55} & \ding{51} & 0.8576 & 0.7187 \\
\ding{51} & \ding{51} & \ding{51} & \textbf{0.8812} & \textbf{0.7656} \\
\bottomrule
\end{tabular}
\end{table}


\subsection{Ablation Study}

\paragraph{Effect of Different Uncertainty Measures.}
We examine the impact of different uncertainty configurations on the WebShop dataset.
As shown in Table~\ref{tab:uncertainty}, using only one type of uncertainty or a pairwise combination yields partial improvements but does not match the performance of the full model.
We note that when the method relies solely on entropy-based uncertainty, it aligns with the recent approach proposed in~\cite{xie2025unlocking}. However, while entropy effectively captures the overall "sharpness" of the policy distribution, it tends to overlook nuanced ambiguities during decision-making.

In contrast, least confidence and margin metrics offer a more granular perspective by focusing on the competitive relationship between the top-ranked actions. The superior performance of the unified model suggests that these measures are orthogonal and complementary: they collectively identify diverse "zones of uncertainty" that a solitary metric might fail to detect. These findings imply that different uncertainty types play dominant roles at varying stages of the learning process, fostering a more stable and robust exploration strategy through their synergy.

\begin{wrapfigure}[15]{r}{0.5\columnwidth}
\vspace{-7mm}
\centering
\includegraphics[width=1\linewidth]{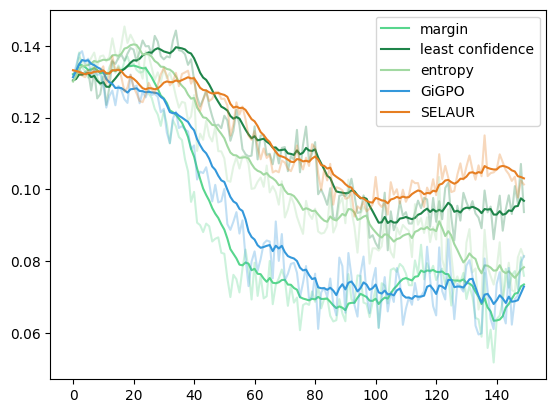}
\vspace{-8mm}
\caption{\small Entropy change during training under different uncertainty strategies.
SELAUR maintains higher entropy throughout training, indicating stronger exploration compared to other methods.}
\label{fig:fail}
\vspace{-20pt} 
\end{wrapfigure}
We further visualize the evolution of entropy during training across different uncertainty configurations, as illustrated in Fig.~\ref{fig:fail}.
Compared to other variants, SELAUR maintains a higher entropy level in the late stage, suggesting that the combined approach promotes greater uncertainty and facilitates broader exploration throughout the learning process.

\begin{wraptable}{r}{0.5\textwidth} 
\vspace{-20pt}
\centering
\caption{\small Comparison of different reward functions on \textit{WebShop} and \textit{AlfWorld}. We report average of success rate(\%) on AlfWorld and WebShop. 
Our proposed SELAUR achieves the best performance, demonstrating a better trade-off between exploration and exploitation.
}
\label{tab:reward-all}

\begin{tabular}{c|cc}
\toprule
 \textbf{Reward Method} & \textbf{AlfWorld} & \textbf{WebShop} \\
\midrule

Baseline         & 0.8828 & 0.6757 \\
Negative         & 0.8516 & 0.7539 \\
Exponential      & 0.8750 & 0.7031 \\
Ours (SELAUR)    & \textbf{0.8906} & \textbf{0.7656} \\

\bottomrule
\end{tabular}
\vspace{-20pt}

\end{wraptable}
\paragraph{Effect of Different Reward Functions} We also explore different ways of incorporating uncertainty into the reward function, as summarized in Table~\ref{tab:reward-all}.
The \textbf{Negative} variant replaces rewards in failure cases with negative values, aiming to penalize uncertain or ineffective actions.
The \textbf{Exponential} variant, on the other hand, attenuates rewards for successful trajectories according to their uncertainty level, encouraging the model to be more confident when success is achieved.
Formally, the exponential formulation is defined as:
\begin{equation}
\tilde r_t^{\text{step}} =
\begin{cases}
0, & \text{if fail}, \\
r_t \cdot \exp(-\hat u^{\text{step}}_t), & \text{otherwise}.
\end{cases}
\end{equation}

Both variants underperform compared to our proposed method.
For the exponential approach, the excessive discounting of rewards likely weakens the reinforcement of successful behaviors, hindering stable policy learning.
For the negative variant, the penalization of uncertainty contradicts our objective of encouraging controlled exploration, thereby limiting the model’s ability to adapt in uncertain states.
In contrast, our method formulation consistently achieves the highest task score and success rate across both benchmarks, effectively balancing exploration and exploitation throughout training.

\subsection{Case Studies}\label{sec:case}

\begin{figure}
    \centering
    \includegraphics[width=1\linewidth]{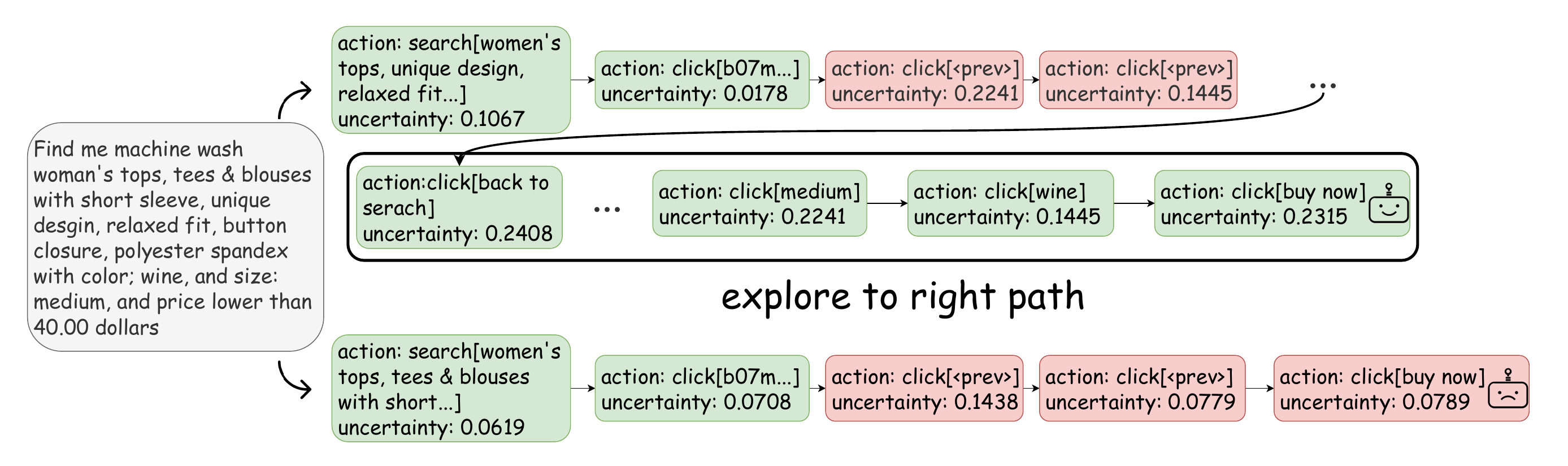}
    \caption{Comparison of action traces in the WebShop environment.
\textbf{Top}: SELAUR leverages uncertainty to explore alternative paths rather than being trapped in a single incorrect trajectory, eventually discovering the correct solution. \textbf{Bottom}: GiGPO exhibits low uncertainty, leading to repetitive and fixed behaviors.
}
    \label{fig:trace}
\vspace{-10 pt}
\end{figure}

To further understand the behavioral differences between GiGPO and SELAUR, we select several representative qualitative trajectories from the WebShop environment. When performing the same product search tasks, our model not only achieves higher success rates but also exhibits more efficient and coherent decision trajectories.
As illustrated in Figure~\ref{fig:trace}, GiGPO often revisits similar intermediate actions repeatedly, showing signs of uncertainty collapse and becoming trapped in local exploration loops. In contrast, although our method may initially encounter similar detours, its uncertainty-aware guidance allows it to dynamically adjust exploration behavior according to confidence levels. This mechanism helps the agent recognize and escape from unproductive cycles after a few repetitions, ultimately steering it toward the correct path.

\section{Conclusion}
\vspace{-5pt}
In this work, we presented SELAUR, a reinforcement learning framework that incorporates multi-type uncertainty signals into reward shaping to more effectively exploit information from both successful and unsuccessful trajectories. By estimating uncertainty at the token, step, and trajectory levels, SELAUR enables principled reward modulation that facilitates deeper exploration, mitigates entrapment in suboptimal reasoning loops, and extracts supervisory signal from trajectories that would otherwise provide limited learning value. Empirical results across multiple benchmarks demonstrate consistent improvements over strong baselines, indicating that uncertainty-aware reward shaping contributes to more stable learning dynamics, higher task success rates, and enhanced adaptability in complex interactive environments.
Future research will investigate the extension of SELAUR to vision-centric and multimodal tasks, as well as evaluate its efficacy in domains characterized by high-dimensional observations and more diverse forms of uncertainty.
\vspace{-10pt}

\section*{Acknowledgments}
The work was partially supported by NSF awards 2442477, 2312862,2533996, NSF-Simons SkAI Institute, NIH R01AG091762, UIC IEHDSR seed funding. We thank Amazon Research Awards, Cisco Research Awards, Google, and OpenAI for providing us with API credits. The authors acknowledge Research Computing at Arizona State University and Johns Hopkins University, NSF ACCESS for providing computing resources.  The views and conclusions in this paper should not be interpreted as representing any funding agencies.
\bibliographystyle{splncs04}
\bibliography{custom}

\end{document}